\documentclass{article}

\usepackage{microtype}
\usepackage{graphicx}
\usepackage{booktabs} %

\usepackage{hyperref}

\usepackage{xcolor}  %

\usepackage[accepted]{icml2019}  %

\usepackage{enumitem}
\usepackage{placeins}
\usepackage{subcaption}
\usepackage{todonotes}  %
\usepackage{xspace}
\usepackage{booktabs} 

\usepackage{amsmath}
\DeclareMathOperator*{\argmax}{arg\,max}

\def\nasbenchnoformat{NAS-Bench-101}
\def\nasbench{\nasbenchnoformat\xspace}
\def\nasbenchmini{NAS-Bench-Mini\xspace}
\def\hpobenchnoformat{NAS-HPO-Bench}
\def\hpobench{\hpobenchnoformat\xspace}
\newcommand{\op}[1]{{\texttt{#1}}}
\newcommand{\citei}[1]{\citeauthor{#1}, \citeyear{#1}}
\def\eg{\textit{e.g.}\xspace}
\def\ie{\textit{i.e.}\xspace}

\newcommand{\tmax}{E_{\mathrm{max}}}
\newcommand{\tstop}{E_{\mathrm{stop}}}

\newcommand{\eat}[1]{} %

\icmltitlerunning{\nasbenchnoformat}

\begin{document}

\twocolumn[
\icmltitle{\nasbenchnoformat: Towards Reproducible Neural Architecture Search}
\icmlsetsymbol{equal}{*}

\begin{icmlauthorlist}
\icmlauthor{Chris Ying}{equal,google}
\icmlauthor{Aaron Klein}{equal,freiburg}
\icmlauthor{Esteban Real}{google}
\icmlauthor{Eric Christiansen}{google}
\icmlauthor{Kevin Murphy}{google}
\icmlauthor{Frank Hutter}{freiburg}
\end{icmlauthorlist}

\icmlaffiliation{freiburg}{Department of Computer Science, University of Freiburg, Germany}
\icmlaffiliation{google}{Google Brain, Mountain View, California, USA}

\icmlcorrespondingauthor{Chris Ying}{\mbox{contact@chrisying.net}}
\icmlcorrespondingauthor{Aaron Klein}{\mbox{kleinaa@cs.uni-freiburg.de}}
\icmlcorrespondingauthor{Esteban Real}{\mbox{ereal@google.com}}

\icmlkeywords{NASBench, NAS, neuro-architecture search, neural architecture search, benchmark, AutoML, automatic machine learning, dataset, reproducible, reproducibility}

\vskip 0.3in
]

\printAffiliationsAndNotice{\icmlEqualContribution} %

\begin{abstract}
Recent advances in neural architecture search (NAS) demand tremendous computational resources, which makes it difficult to reproduce experiments and imposes a barrier-to-entry to researchers without access to large-scale computation.
We aim to ameliorate these problems by introducing NAS-Bench-101, the first public architecture dataset for NAS research.
To build NAS-Bench-101, we carefully constructed a compact, yet expressive, search space, exploiting graph isomorphisms to identify 423k unique convolutional architectures.
We trained and evaluated all of these architectures multiple times on CIFAR-10 and compiled the results into a large dataset of over 5 million trained models.
This allows researchers to evaluate the quality of a diverse range of models in milliseconds by querying the pre-computed dataset.
We demonstrate its utility by analyzing the dataset as a whole and by benchmarking a range of architecture optimization algorithms.
\end{abstract}

\section{Introduction}

Many successes in deep learning \cite{krizhevsky2012imagenet,goodfellow2014generative,sutskever2014sequence} have resulted from novel neural network architecture designs. For example, in the field of image classification, research has produced numerous ways of combining neural network layers into unique architectures, such as Inception modules \cite{szegedy2015going}, residual connections \cite{he2016deep}, or dense connections \cite{huang2016densely}.
This proliferation of choices has fueled research into \textit{neural architecture search (NAS)}, which casts the discovery of new architectures as an optimization problem \cite{baker2016designing,zoph2016neural,real2017large,elsken-jmlr19a}.
This has resulted in state of the art performance in the domain of image classification \cite{zoph2017learning,real2018regularized,huang2018gpipe},
and has shown promising results in other domains, such as sequence modeling \cite{zoph2016neural, So2019TheET}.

Unfortunately, NAS research is notoriously hard to reproduce \cite{2019arXiv190207638L,DBLP:journals/corr/abs-1902-08142}.
First, some methods require months of compute time (\eg, \citei{zoph2017learning}), making these methods inaccessible to most researchers.
Second, while recent improvements \cite{liu2017progressive,pham2018efficient,liu2018darts} have yielded more efficient methods, different methods are not comparable to each other due to different training procedures and different search spaces, which make it difficult to attribute the success of each method to the search algorithm itself.

To address the issues above, this paper introduces \nasbench, the first architecture-dataset for NAS. 
To build it, we trained and evaluated a large number of different convolutional neural network (CNN) architectures on CIFAR-10~\cite{krizhevsky2009learning}, utilizing over 100 TPU years of computation time. 
We compiled the results into a large table which maps 423k unique architectures to metrics including run time and accuracy.
This enables NAS experiments to be run via querying a table instead of performing the usual costly train and evaluate procedure.
Moreover, the data, search space, and training code is fully public \footnote{
Data and code for NAS-Bench-101 available at \url{https://github.com/google-research/nasbench}.}, to foster reproducibility in the NAS community.

Because \nasbench exhaustively evaluates a search space, it permits, for the first time, a comprehensive \mbox{analysis} of a NAS search space as a whole.
We illustrate such potential by measuring search space properties relevant to architecture search.
Finally, we demonstrate its application to the analysis of algorithms by benchmarking a wide range of open source architecture/hyperparameter search methods, including evolutionary approaches, random search, and Bayesian optimization.

In summary, our contributions are the following:
\begin{itemize}[noitemsep,topsep=-5pt,leftmargin=*]
    \item We introduce \nasbench, the first large-scale, open-source architecture dataset for NAS (Section~\ref{sec:dataset});
    \item We illustrate how to use the dataset to analyze the nature of the search space, revealing insights which may guide the design of NAS algorithms (Section~\ref{sec:nas-dataset});
    \item We illustrate how to use the dataset to perform fast benchmarking of various open-source NAS optimization algorithms  (Section~\ref{sec:nas-benchmark}).
\end{itemize}

\section{The NASBench Dataset}
\label{sec:dataset}

The \nasbench dataset is a table which maps neural network architectures to their training and evaluation metrics.
Most NAS approaches to date have trained models on the CIFAR-10 classification set because its small images allow relatively fast neural network training. 
Furthermore, models which perform well on CIFAR-10 tend to perform well on harder benchmarks, such as ImageNet \cite{krizhevsky2012imagenet} when scaled up \cite{zoph2017learning}).
For these reasons, we also use CNN training on CIFAR-10 as the basis of \nasbench.

\subsection{Architectures}
\label{sec:dataset_archs}

\begin{figure}[h!]
\sbox0{\begin{subfigure}[c]{0.45\linewidth}
    \centering
    \includegraphics[width=0.9\linewidth]{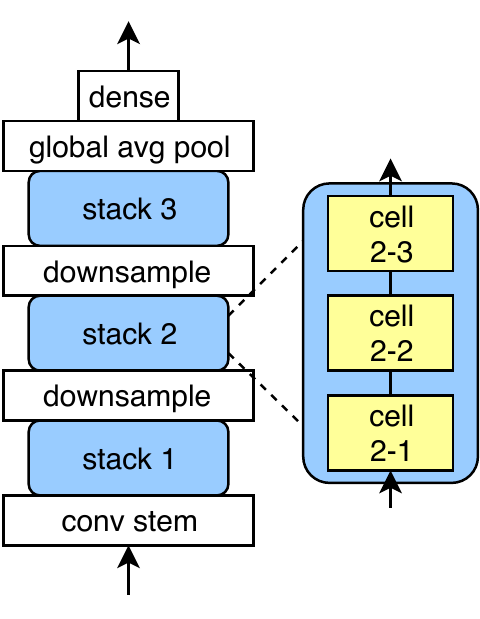}
\end{subfigure}}
\sbox1{\begin{subfigure}[c]{0.45\linewidth}
    \centering
    \includegraphics[width=0.9\linewidth, trim={0 0 0 0}]{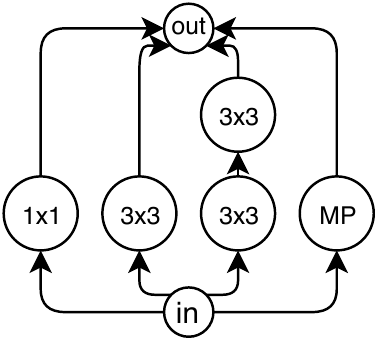}
\end{subfigure}}
\sbox2{\begin{subfigure}[c]{0.45\linewidth}
    \centering
    \includegraphics[width=0.6\linewidth]{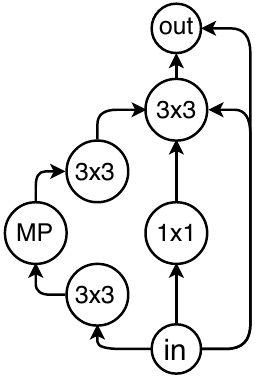}
\end{subfigure}}
\sbox3{\begin{subfigure}[c]{0.45\linewidth}
    \centering
    \includegraphics[width=0.95\linewidth]{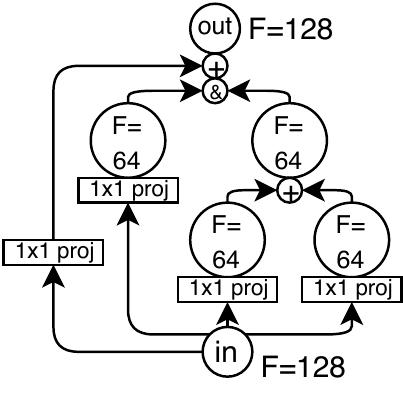}
\end{subfigure}}
\begin{center}
\begin{tabular}{cc}
\usebox0 & \usebox1\\
\usebox2 & \usebox3\\
\end{tabular}
\end{center}
\caption{(top-left) The outer skeleton of each model. (top-right) An Inception-like cell with the original 5x5 convolution approximated by two 3x3 convolutions (concatenation and projection operations omitted). (bottom-left) The cell that attained the lowest mean test error (projection layers omitted). (bottom-right) An example cell that demonstrates how channel counts are automatically determined (``+'' denotes addition and ``\&'' denotes concatenation; $1 \times 1$ projections are used to scale channel counts).}
\label{arch_figures}
\end{figure}

Similar to other NAS approaches, we restrict our search for neural net topologies to the space of small feedforward structures, usually called \textit{cells}, which we describe below. We stack each cell 3 times, followed by a downsampling layer, in which the image height and width are halved via max-pooling and the channel count is doubled.
We repeat this pattern 3 times, followed by global average pooling and a final dense softmax layer.
The initial layer of the model is a \textit{stem} consisting of one $3 \times 3$ convolution with $128$ output channels.
See Figure~\ref{arch_figures}, top-left, for an illustration of the overall network structure.
Note that having a stem followed by stacks of cells is a common pattern both in hand-designed image classifiers \cite{he2016deep,huang2016densely,hu2017squeeze} and in NAS search spaces for image classification.
Thus, the variation in the architectures arises from variation in the cells.

The space of cell architectures consists of all possible directed acyclic graphs on $V$ nodes, where each possible node has one of $L$ labels, representing the corresponding operation.
Two of the vertices are specially labeled as operation \op{IN} and \op{OUT}, representing the input and output tensors to the cell, respectively. 
Unfortunately, this space of labeled DAGs grows exponentially in both $V$ and $L$.
In order to limit the size of the space to allow exhaustive enumeration, we impose the following constraints:
\begin{itemize}[noitemsep,topsep=-5pt,leftmargin=*]
    \item We set $L=3$, using only the following operations:
        \begin{itemize}[noitemsep,topsep=-5pt,leftmargin=*]
            \item $3 \times 3$ convolution
            \item $1 \times 1$ convolution
            \item $3 \times 3$ max-pool
        \end{itemize}
    \item We limit $V \leq 7$.
    \item We limit the maximum number of edges to $9$.
\end{itemize}
All convolutions utilize batch normalization followed by ReLU.
These constraints were chosen to ensure that the search space still contains ResNet-like and Inception-like cells \cite{he2016deep,szegedy2016rethinking}.
An example of an Inception-like cell is illustrated in Figure~\ref{arch_figures}, top-right.
We intentionally use convolutions instead of separable convolutions to match the original designs of ResNet and Inception, although this comes as the cost of being more parameter-heavy than some of the more recent state-of-the-art architectures like AmoebaNet \cite{real2018regularized}.

\subsection{Cell encoding}
\label{sec:encoding}

There are multiple ways to encode a cell and different encodings may favor certain algorithms by biasing the search space.
For most of our experiments, we chose to use a very general encoding: a 7-vertex directed acyclic graph, represented by a $7 \times 7$ upper-triangular binary matrix, and a list of 5 labels, one for each of the 5 intermediate vertices (recall that the input and output vertices are fixed)
Since there are 21 possible edges in the matrix and 3 possible operations for each label, there are $2^{21} * 3^{5} \approx 510M$ total unique models in this encoding.
In Supplement~\ref{sec:supp_encoding}, we also discuss an alternative encoding.

However, a large number of models in this space are invalid (i.e., there is no path from the input vertex, or the number of total edges exceeds 9). 
Furthermore, different graphs in this encoding may not be computationally unique.
The method which we used to identify and enumerate unique graphs is described in Supplement~\ref{sup:deduping_supplement}.
After de-duplication, there are approximately 423k unique graphs in the search space.

\subsection{Combine semantics}
\label{sec:combine}

Translating from the graph to the corresponding neural network is straightforward, with one exception.
When multiple edges point to the same vertex, the incoming tensors must be combined.
Adding them or concatenating them are both standard techniques. To support both ResNet and Inception-like cells and to keep the space tractable, we adopted the following fixed rule: tensors going to the output vertex are concatenated and those going into other vertices are summed.
The output tensors from the input vertex are projected in order to match the expected input channel counts of the subsequent operations.
This is illustrated in Figure~\ref{arch_figures}, bottom-right.

\subsection{Training}
\label{sec:dataset_training}

The training procedure forms an important part of an architecture search benchmark, since different training procedures can lead to very substantial performance differences.
To counter this issue and allow comparisons of NAS algorithms on equal grounds, we designed and open-sourced a single general training pipeline for all models in the dataset.

\textbf{Choice of hyperparameters.} We utilize a single, fixed set of hyperparameters for all \nasbench models.
This set of hyperparameters was chosen to be robust across different architectures by performing a coarse grid search optimizing the average accuracy of a set of 50 randomly-sampled architectures from the space.
This is similar to standard practice in the literature \cite{zoph2017learning,liu2017progressive,real2018regularized} and is further justified by our experimental analysis in Section~\ref{hparams_section}.

\textbf{Implementation details.} All models are trained and evaluated on CIFAR-10 (40k training examples, 10k validation examples, 10k testing examples), using standard data augmentation techniques \cite{he2016deep}.
The learning rate is annealed via cosine decay \cite{loshchilov2016sgdr} to 0 in order to reduce the variance between multiple independent training runs.
Training is performed via RMSProp \cite{tieleman2012lecture} on the cross-entropy loss with L2 weight decay.
All models were trained on the TPU v2 accelerator.
The code, implemented in TensorFlow, along with all chosen hyperparameters, is publicly available at \url{https://github.com/google-research/nasbench}.

\textbf{3 repeats and 4 epoch budgets.}
We repeat the training and evaluation of all architectures 3 times to obtain a measure of variance.
Also, in order to allow the evaluation of multi-fidelity optimization methods, \eg, Hyperband \cite{li2017hyperband}), we trained all our architectures with four increasing epoch budgets: $\tstop \in \{\tmax/3^3, \tmax/3^2, \tmax/3, \tmax\} = \{4, 12, 36, 108\}$ epochs.
In each case, the learning rate is annealed to 0 by epoch $\tstop$.\footnote{
Instead of 4 epoch budgets, we could have trained single long runs and used the performance at intermediate checkpoints as benchmarking data for early stopping algorithms. However, because of the learning rate schedule, such checkpoints would have occurred when the learning rates are still high, leading to noisy accuracies that do not correlate well with the final performance.}
We thus trained $3 \times 423k \sim 1.27M$ models for each value of $\tstop$, and thus $4 \times 1.27M \sim 5M$ models overall.

\subsection{Metrics}
\label{sec:dataset_metrics}

We evaluated each architecture $A$ after training three times with different random initializations, and did this for each of the $4$ budgets $\tstop$ above. As a result, the dataset is a mapping from the $(A, \tstop, \mathtt{trial\#})$ to the 
following quantities:
\begin{itemize}[noitemsep,topsep=-5pt,leftmargin=*]
    \item training accuracy;
    \item validation accuracy;
    \item testing accuracy;
    \item training time in seconds; and
    \item number of trainable model parameters.
\end{itemize}
\vspace{5pt}

Only metrics on the training and validation set should be used to search models within a single NAS algorithm, and testing accuracy should only be used for an offline evaluation.
The training time metric allows benchmarking algorithms that optimize for accuracy while operating under a time limit (Section~\ref{sec:nas-benchmark}) and also allows the evaluation of multi-objective optimization methods.
Other metrics that do not require retraining can be computed using the released code.

\subsection{Benchmarking methods}
\label{sec:dataset_bencharking_methods}

One of the central purposes of the dataset is to facilitate benchmarking of NAS algorithms.
This section establishes recommended best practices for using \nasbench which we followed in our subsequent analysis; we also refer to Supplement~\ref{sec:guidelines} for a full set of best practices in benchmarking with NAS-Bench-101.

The goal of NAS algorithms is to find architectures that have high testing accuracy at epoch $\tmax$.
To do this, we repeatedly query the dataset at $(A, \tstop)$ pairs, where $A$ is an architecture in the search space and $\tstop$ is an allowed number of epochs ($\tstop \in \{4, 12, 36, 108\}$). Each query does a look-up using a random trial index, drawn uniformly at random from $\{1,2,3\}$, to simulate the stochasticity of SGD training.

While searching, we keep track of the best architecture $\hat{A}_i$ the algorithm has found after each function evaluation $i$, as ranked by its \textit{validation accuracy}.
To best simulate real world computational constratints, we stop the search run when the total ``training time'' exceeds a fixed limit.
After each complete search rollout, we query the corresponding mean \textit{test accuracy} $f(\hat{A}_i)$ for that model (test accuracy should never be used to guide the search itself).
Then we compute the immediate test regret: $r(\hat{A}_i) = f(\hat{A}_i) - f(A^*)$, where $A^*$ denotes the model with the highest mean test accuracy in the entire dataset. This regret becomes the score for the search run.
To measure the robustness of different search algorithms, a large number of independent search rollouts should be conducted.

\section{NASBench as a Dataset}
\label{sec:nas-dataset}

In this section, we analyze the \nasbench dataset as a whole to gain some insight into the role of neural network operations and cell topology in the performance of convolutional neural networks. In doing so, we hope to shed light on the loss landscape that is traversed by NAS algorithms.

\subsection{Dataset statistics}
\label{sec:analysis_statistics}
\begin{figure}[t]
\begin{center}
 \includegraphics[width=0.49\columnwidth]{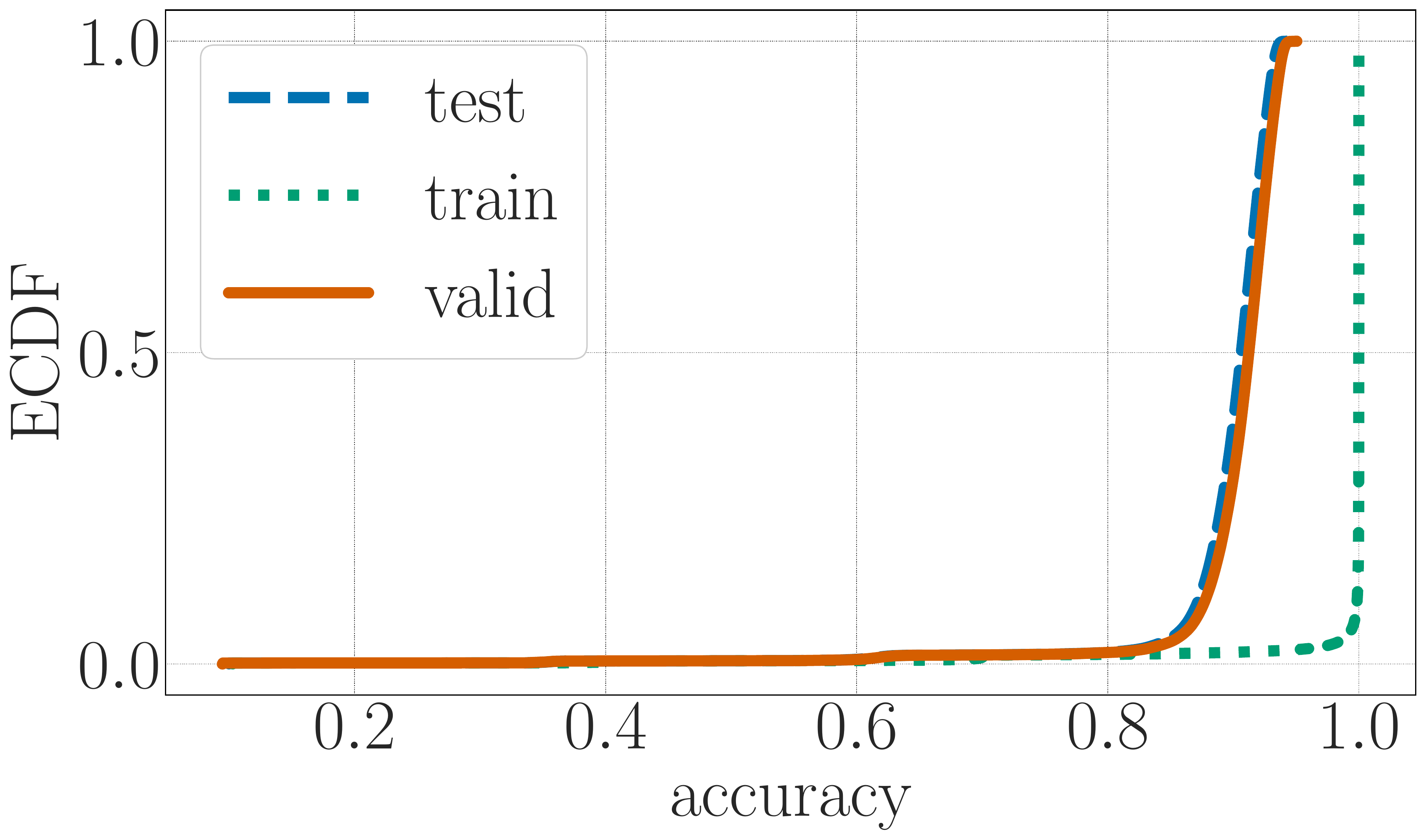}
 \includegraphics[width=0.49\columnwidth]{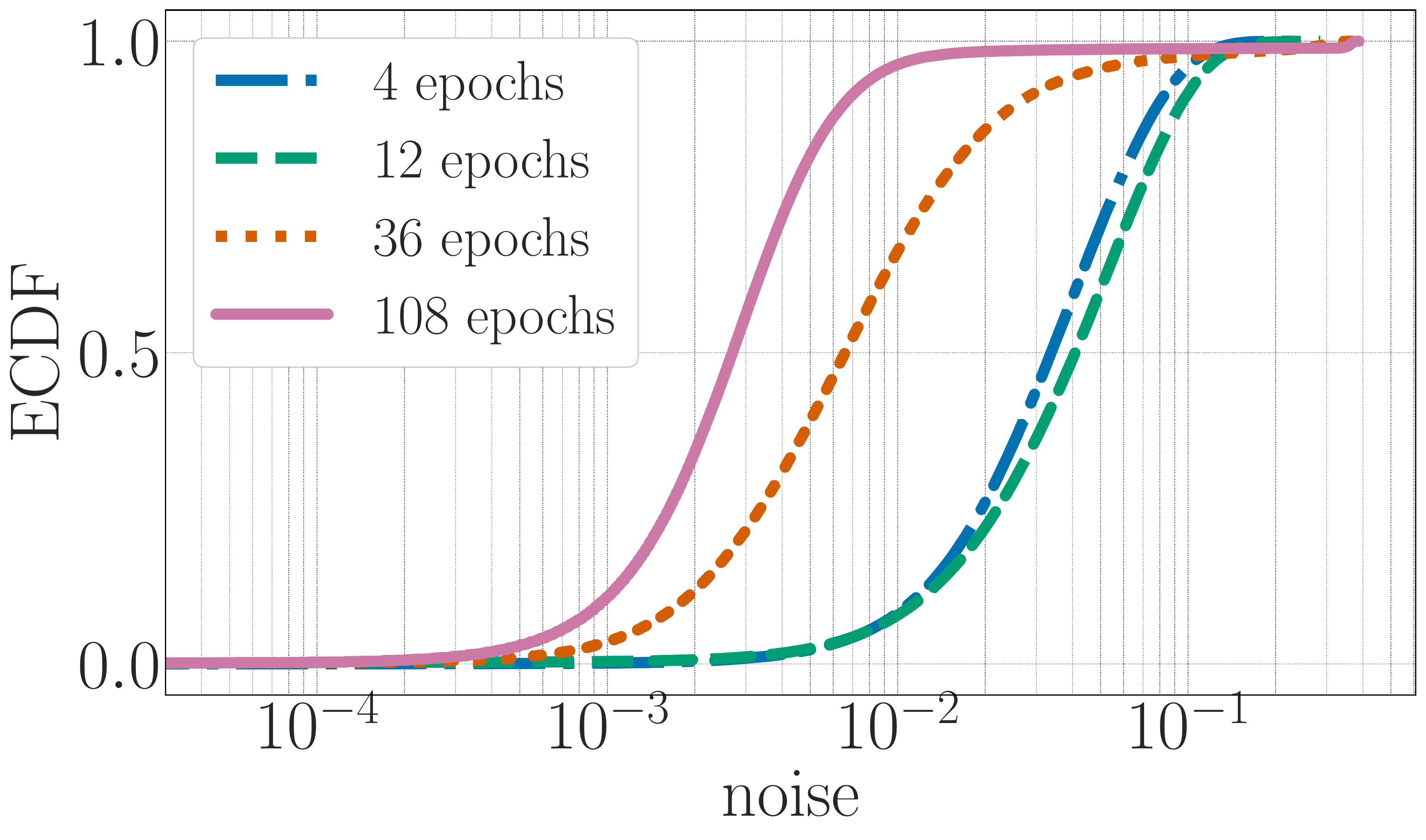}
 \caption{The empirical cumulative distribution (ECDF) of all valid configurations for: (left) the train/valid/test accuracy after training for 108 epochs and (right) the noise, defined as the standard deviation of the test accuracy between the three trials, after training for 12, 36 and 108 epochs.}
 \label{fig:ecdf_plots}
\end{center}
\end{figure}

First we study the empirical cumulative distribution (ECDF) of various metrics across all architectures in Figure \ref{fig:ecdf_plots}.
Most of the architectures converge and reach $100\%$ training accuracy.
The validation accuracy and test accuracy are both above $90\%$ for a majority of models.
The best architecture in our dataset (Figure \ref{arch_figures}) achieved a mean test accuracy of $94.32\%$.
For comparison, the ResNet-like and Inception-like cells attained $93.12\%$ and $92.95\%$, respectively, which is roughly in-line with the performance of the original ResNet-56 ($93.03\%$) on CIFAR-10 \cite{he2016deep}.
We observed that the correlation between validation and test accuracy is extremely high ($r=0.999$) at 108 epochs which suggests that strong optimizers are unlikely to overfit on the validation error. Due to the stochastic nature of the training process, training and evaluating the same architecture will generally lead to a small amount of noise in the accuracy.
We also observe, as expected, that the noise between runs is lower at longer training epochs.

\begin{figure}[t]
\begin{center}
 \includegraphics[width=0.49\linewidth]{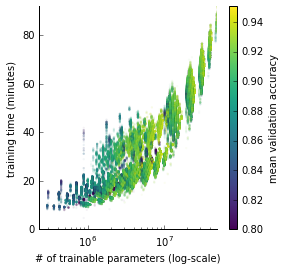}
 \includegraphics[width=0.49\linewidth]{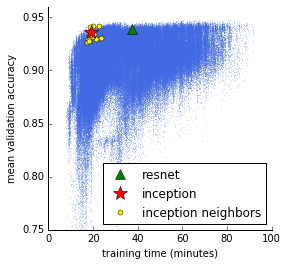}
 \caption{(left) Training time vs. trainable parameters, color-coded by validation accuracy. (right) Validation accuracy vs. training time with select cell architectures highlighted. Inception neighbors are the graphs which are 1-edit distance away from the Inception cell.
 }
 \label{fig:scatter_plots}
\end{center}
\end{figure}

Figure \ref{fig:scatter_plots} investigates the relationship between the number of parameters, training time, and validation accuracy of models in the dataset.
The left plot suggests that there is positive correlation between all of these quantities.
However parameter count and training time are not the only factors since the best cell in the dataset is not the most computationally intensive one.
Hand-designed cells, such as ResNet and Inception, perform near the Pareto frontier of accuracy over cost, which suggests that topology and operation selection are critical for finding both high-accuracy and low-cost models.

\subsection{Architectural design}
\begin{figure}[t]
\begin{center}
 \includegraphics[width=\linewidth]{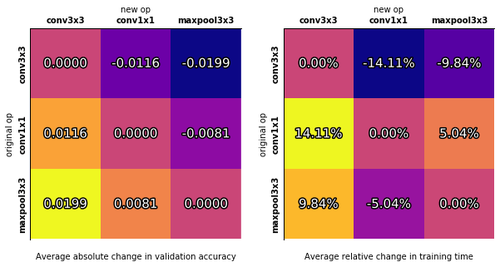}
 \caption{Measuring the aggregated impact of replacing one operation with another on (left) absolute validation accuracy and (right) relative training time.}
 \label{fig:op_change_matrix}
\end{center}
\end{figure}

\nasbench presents us with the unique opportunity to investigate the impact of various architectural choices on the performance of the network.
In Figure \ref{fig:op_change_matrix}, we study the effect of replacing each of the operations in a cell with a different operation.
Not surprisingly, replacing a $3\times 3$ convolution with a $1\times 1$ convolution or $3\times 3$ max-pooling operation generally leads to a drop in absolute final validation accuracy by $1.16\%$ and $1.99\%$, respectively.
This is also reflected in the relative change in training time, which decreases by $14.11\%$ and $9.84\%$.
Even though $3\times 3$ max-pooling is parameter-free, it appears to be on average $5.04\%$ more expensive in training time than $1\times 1$ convolution and also has an average absolute validation accuracy $0.81\%$ lower.
However, some of the top cells in the space (ranked by mean test accuracy, i.e., Figure \ref{arch_figures}) contain max-pool operations, so other factors must also be at play and replacing all $3\times 3$ max-pooling operations with $1\times 1$ convolutions is not necessarily a globally optimal choice.

\begin{figure}[t]
\begin{center}
 \includegraphics[width=\linewidth]{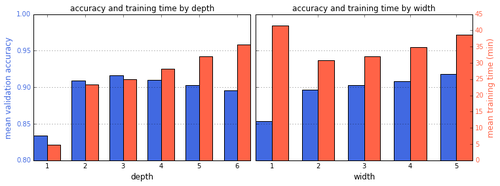}
 \caption{Comparing mean validation accuracy and training time for cells by (left) depth, measured by length of longest path from inpu to output, and (right) width, measured by maximum directed cut on the graph.}
 \label{fig:depth_width}
\end{center}
\end{figure}

In Figure \ref{fig:depth_width}, we also investigate the role of depth vs. width.
In terms of average validation accuracy, it appears that a depth of 3 is optimal whereas increasing width seems to increase the validation accuracy up to 5, the maximum width of networks in the dataset.
The training time of networks increases as networks get deeper and wider with one exception: width 1 networks are the most expensive.
This is a consequence of the combine semantics (see Section \ref{sec:combine}), which skews the training time distributions because all width 1 networks are simple feed-forward networks with no branching, and thus the activation maps are never split via their channel dimension.

\subsection{Locality}

\begin{figure}[h!]
\begin{center}
 \includegraphics[width=0.49\linewidth]{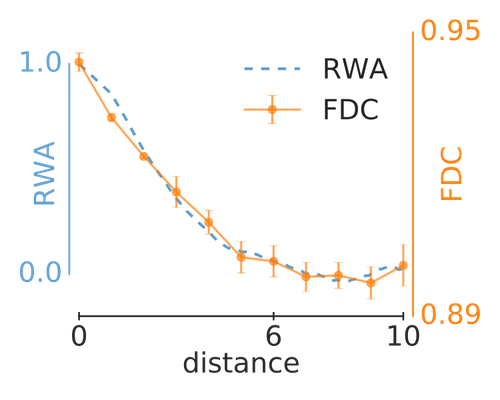}
 \includegraphics[width=0.49\linewidth]{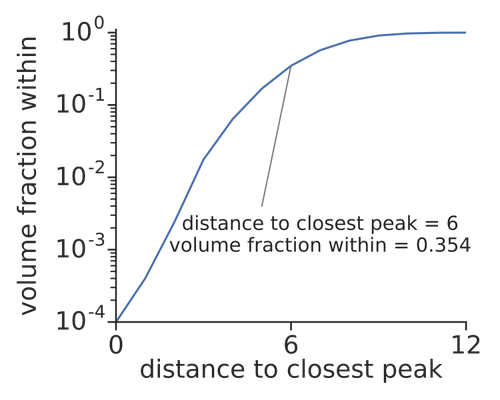}
 \caption{(left) RWA for the full space and the FDC relative to the global maximum. To plot both curves on a common horizontal axis, the autocorrelation curve is drawn as a function of the square root of the autocorrelation shift, to account for the fact that a random walk reaches a mean distance $\sqrt{N}$ after $N$ steps. (right) Fraction of the search space volume that lies within a given distance to the closest high peak.}
 \label{fig:spikiness}
\end{center}
\end{figure}

NASBench exhibits \textit{locality}, a property by which architectures that are ``close by'' tend to have similar performance metrics. This property is exploited by many search algorithms. We define ``closeness'' in terms of \textit{edit-distance}: the smallest number of changes required to turn one architecture into another; one \textit{change} entails flipping the operation at a vertex or the presence/absence of an edge. A popular measure of locality is the random-walk autocorrelation (RWA), defined as the autocorrelation of the accuracies of points visited as we perform a long walk of random changes through the space \cite{weinberger1990correlated,stadler1996landscapes}. The RWA (Figure~\ref{fig:spikiness}, left) shows high correlations for lower distances, indicating locality. The correlations become indistinguishable from noise beyond a distance of about 6.

While the RWA aggregates across the whole space, we can also consider regions of particular interest. For example, Figure~\ref{fig:scatter_plots} (right) displays the neighbors of the Inception-like cell, indicating a degree of locality too, especially in terms of accuracy. Another interesting region is that around a global accuracy maximum. To measure locality within this neighborhood, we used the fitness-distance correlation metric (FDC, \citet{jones1995evolutionary}). Figure~\ref{fig:spikiness} (left) shows that there is locality around the global maximum as well and the peak also has a coarse-grained width of about 6. 

More broadly, we can consider how rare it is to be near a global maximum. In the cell encoding described in Section \ref{sec:encoding}, the best architecture (\ie, the one with the highest mean testing accuracy) has 4 graph isomorphisms, producing 4 distinct peaks in our encoded search space. Moreover there are 11 other architectures whose mean test accuracy is within 2 times standard error of the mean of the best graph. Including the isomorphisms of these, too, there are 11\,570 points in the 510M-point search space corresponding to these \textit{top graphs}, meaning that the chance of hitting one of them with a random sample is about 1 to 50000. Figure~\ref{fig:spikiness} (right) shows how much volume of the search space lies near these graphs; in particular, 35.4\% of the search space is within a distance of 6 from the closest top graph. Since the basin of attraction for local search appears to have a width of about 6, this suggests that locality-based search may be a good choice for this space. 

\section{NASBench as a Benchmark}
\label{sec:nas-benchmark}

\subsection{Comparing NAS algorithms}

In this section we establish baselines for future work by using our dataset to compare some popular algorithms for which open source code is available.
Note that the intention is not to answer the question ``\emph{Which methods work best on this benchmark?}'', but rather to demonstrate the utility of a reproducible baseline.

We benchmarked a small set of NAS and hyperparameter optimization (HPO) algorithms with publicly available implementations: random search (RS)~\cite{bergstra2012random}, regularized evolution (RE)~\cite{real2018regularized}, SMAC~\cite{hutter2011sequential}, TPE~\cite{bergstra2011algorithms}, Hyperband (HB)~\citep{li2017hyperband}, and BOHB~\cite{falkner2018bohb}. We follow the guidelines established in Section~\ref{sec:dataset_bencharking_methods}.
Due to its recent success for NAS~\citep{zoph2016neural}, we also include our own implementation of reinforcement learning (RL) as an additional baseline, since an official implementation is not available. However, instead of using an LSTM controller, which we found to perform worse, we used a categorical distribution for each parameter and optimized the probability values directly with REINFORCE.
Supplement~\ref{sup:implementation_details} has additional implementation details for all methods.

NAS algorithms based on weight sharing~\cite{pham2018efficient,liu2018darts}
or network morphisms~\cite{cai2017reinforcement,elsken2018multi}
cannot be directly evaluated on the dataset, so we did not include them.
We also do not include Gaussian process--based HPO methods~\citep{ShahriariSWAF16}, such as Spearmint~\citep{snoek2012practical}, since they tend to have problems in high-dimensional discrete optimization tasks~\cite{EggFeuBerSnoHooHutLey13}.
While Bayesian optimization methods based on Bayesian neural networks~\citep{snoek-icml15,springenberg-nips16} are generally applicable to this benchmark, we found their computational overhead compared to the other methods to be prohibitively expensive for an exhaustive empirical evaluation.
The benchmarking scripts we used are publicly available\footnote{ \url{https://github.com/automl/nas\_benchmarks}}.
For all optimizers we investigate their own main meta-parameters in Supplement~\ref{sec:supp_hyperopt} (except for TPE where the open-source implementation does not allow to change the meta-parameters) and report here the performance based on the best found settings. 

\begin{figure*}[t]
\begin{center}
  \includegraphics[width=0.45\linewidth]{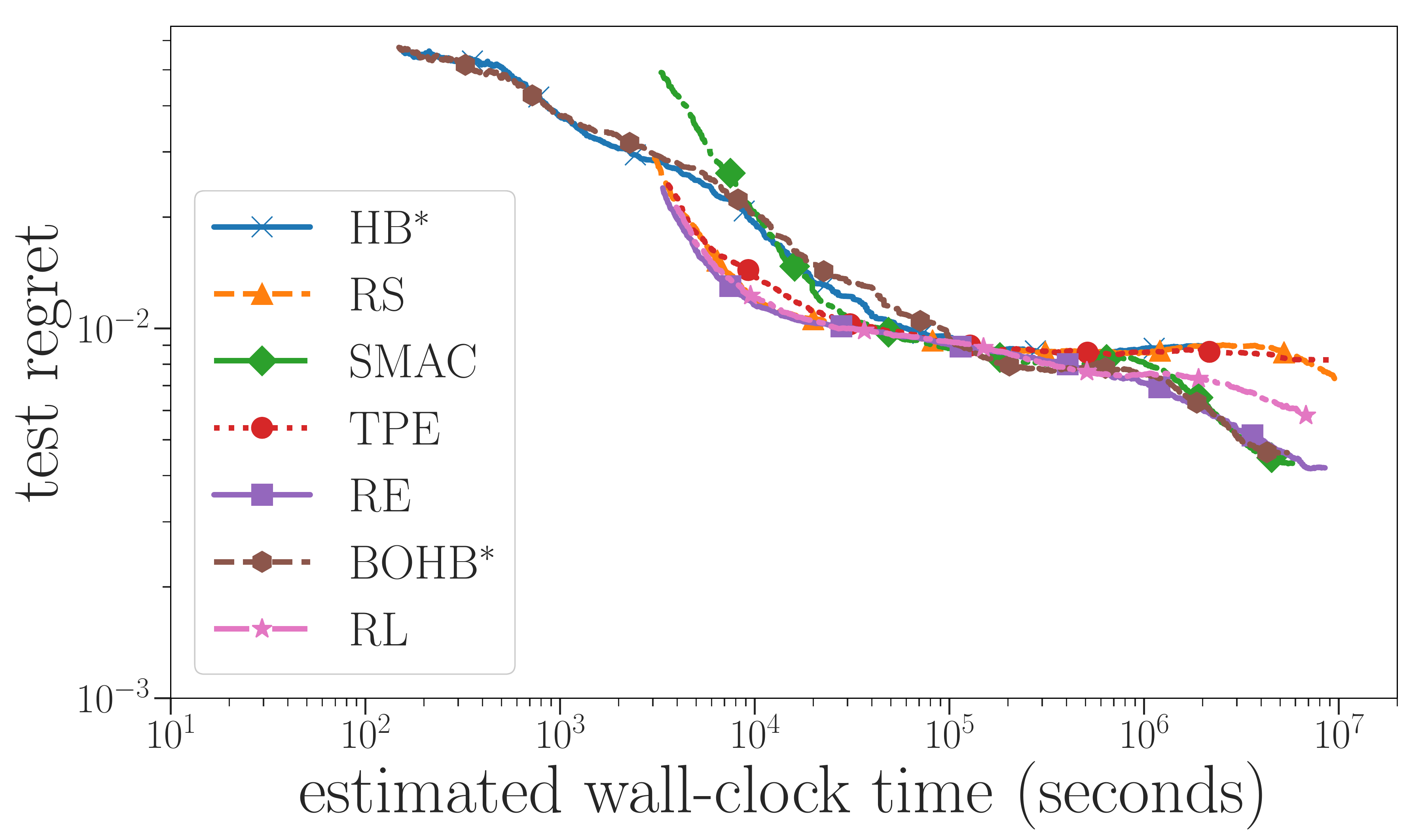}
  \includegraphics[width=0.45\linewidth]{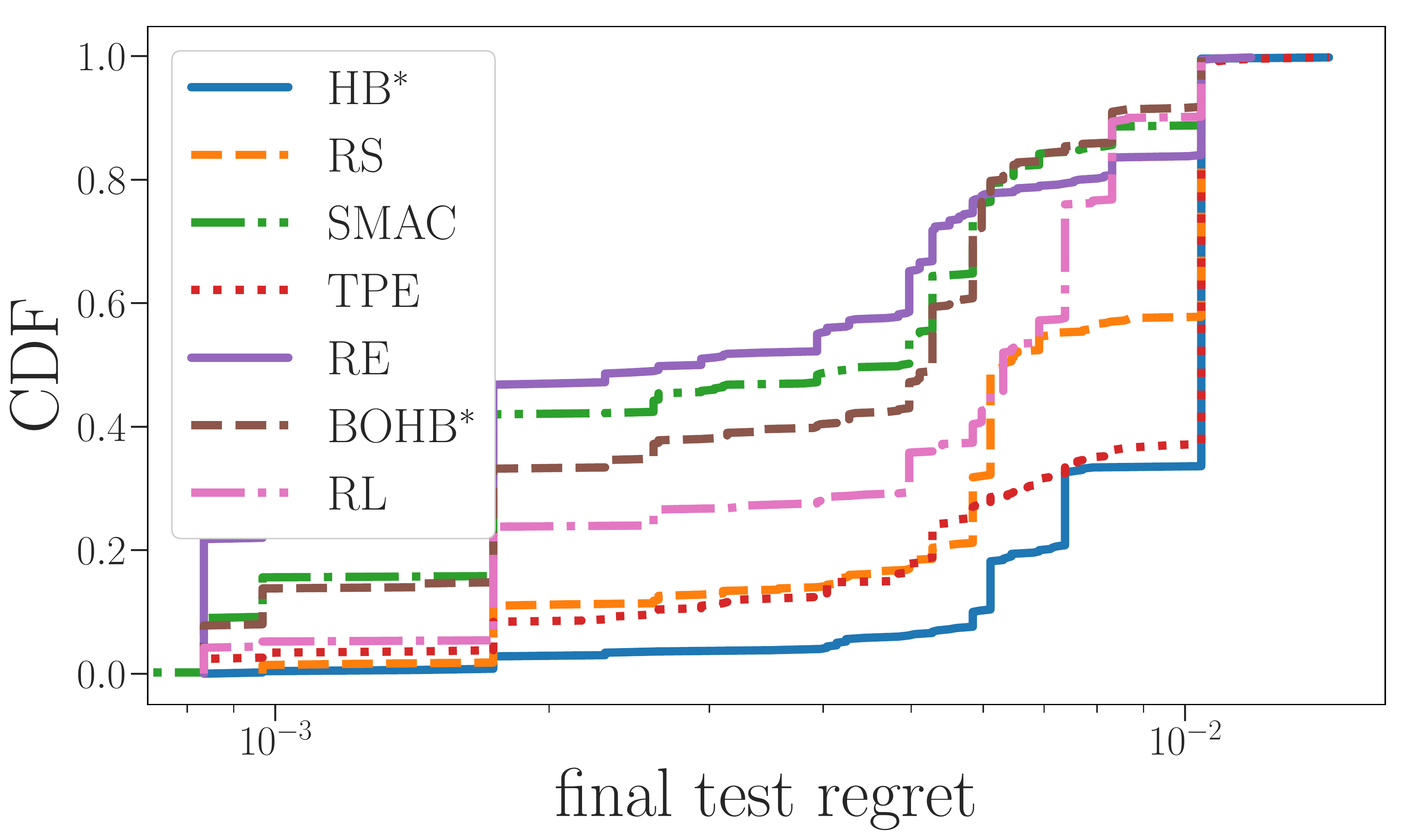}
  \caption{(left) Comparison of the performance of various search algorithms. The plot shows the mean performance of 500 independent runs as a function of the estimated training time. (right) Robustness of different optimization methods with respect to the seed for the random number generator. *HB and BO-HB are budget-aware algorithms which query the dataset a shorter epoch lengths. The remaining methods only query the dataset at the longest length (108 epochs).
  }
 \label{fig:comparison}
\end{center}
\end{figure*}

Figure~\ref{fig:comparison} (left) shows the mean performance of each of these NAS/HPO algorithms across 500 independent trials. 
The x-axis shows \emph{estimated wall-clock time}, counting the evaluation of each architecture with the time that the corresponding training run took. Note that the evaluation of 500 trials of each NAS algorithm (for up to 10M simulated TPU seconds, i.e., 115 TPU days each) was only made possible by virtue of our tabular benchmark; without it, they would have amounted to over 900 TPU years of computation.

We make the following observations:
\begin{itemize}[noitemsep,topsep=-5pt,leftmargin=*]
    \item RE, BOHB, and SMAC perform best and start to outperform RS after roughly 50\,000 TPU seconds (the equivalent of roughly 25 evaluated architectures); they achieved the final performance of RS about 5 times faster and continued to improve beyond this point.
    \item SMAC, as a Bayesian optimization method, performs this well despite the issue of invalid architectures; we believe that this is due to its robust random forest model. SMAC is slightly slower in the beginning of the search; we assume that this is due to its internal incumbent estimation procedure (which evaluates the same architecture multiple times).
    \item The other Bayesian optimization method, TPE, struggles with this benchmark, with its performance falling back to random search.
    \item The multi-fidelity optimization algorithms HB and BO-HB do not yield the speedups frequently observed compared to RS or Bayesian optimization. We attribute this to the relatively low rank-correlation between the performance obtained with different budgets (see Figure \ref{fig:rank_correlation} in Supplement~\ref{sup:implementation_details}).    
    \item BOHB achieves the same test regret as SMAC and RE after recovering from misleading early evaluations; we attribute this to the fact, that, compared to TPE, it uses a multivariate instead of a univariate kernel density estimator.
    \item Even though RL starts outperforming RS at roughly the same time as the other methods, it converges much slower towards the global optimum.
\end{itemize}

Besides achieving good performance, we argue that robustness, \ie, how sensitive an optimizer is to the randomness in both the search algorithm and the training process, plays an important role in practice for HPO and NAS methods.
This aspect has been neglected in the NAS literature due to the extreme cost of performing many repeated runs of NAS experiments, but with \nasbench performing many repeats becomes trivial. Figure~\ref{fig:comparison} (right) shows the empirical cumulative distribution of the regret after 10M seconds across all 500 runs of each method. For all methods, the final test regrets ranged over roughly an order of magnitude, with RE, BOHB, and SMAC showing the most robust performance.

\subsection{Generalization bootstrap}
\label{sec:analysis_generalization}
To test the generalization of our findings on the dataset, we ideally would need to run the benchmarked algorithms on a larger space of architectures.
However, due to computational limitations, it is infeasible for us to run a large number of NAS trials on a meaningfully larger space.
Instead, to provide some preliminary evidence of generalization, we perform a bootstrapped experiment: we set aside a subset of \nasbench, dubbed \nasbenchmini, and compare the outcomes of algorithms run on \nasbenchmini compared to the full \nasbench.
\nasbenchmini contains all cells within the search space that utilize $6$ or fewer vertices ($64.5k$ unique cells), compared to the full \nasbench that uses up to $7$ vertices ($423k$ unique cells).

\begin{figure}[t]
\begin{center}
 \includegraphics[width=\linewidth]{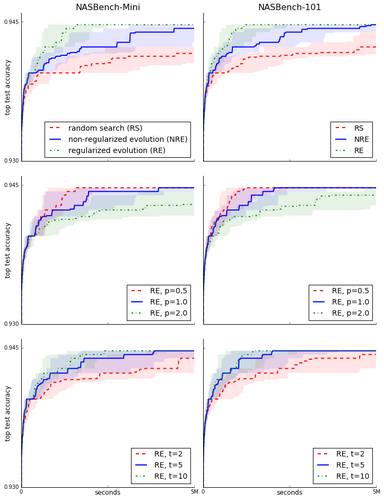}
 \caption{Generalization bootstrap experiments. Each line marks the median of 100 runs and the shaded region is the 25\% to 75\% interquartile range. (top) Comparing random search (RS), non-regularized evolution (NRE), and regularized evolution (RE) against \nasbenchmini and \nasbench. (middle) Comparing RE runs with different mutation rates. (bottom) Comparing RE runs with different tournament sizes.}
 \label{fig:generalization}
\end{center}
\end{figure}

We compare two very similar algorithms (regularized evolution, RE, and non-regularized evolution, NRE) to a baseline (random search, RS).
RE and NRE are identical except that RE removes the \textit{oldest} individual in a population to maintain the population size whereas NRE removes the \textit{lowest fitness} individual.
Figure \ref{fig:generalization} (top) shows the comparison on \nasbenchmini and \nasbench on 100 trials of each algorithm to a fixed time budget.
The plots show that the rankings of the three algorithms (RS $<$ NRE $<$ RE) are consistent across the smaller dataset and the larger one.
Furthermore, we demonstrate that \nasbenchmini can generalize to \nasbench for different hyperparameter settings of a single algorithm (regularized evolution) in Figure \ref{fig:generalization} (middle, bottom).
This suggests that conclusions drawn from \nasbench may generalize to larger search spaces.

\section{Discussion}
\label{sec:discussion}

In this section, we discuss some of the choices we made when designing \nasbench.

\subsection{Relationship to hyperparameter optimization}
\label{hparams_section}

\begin{figure}[t]
\begin{center}
 \includegraphics[width=0.95\linewidth]{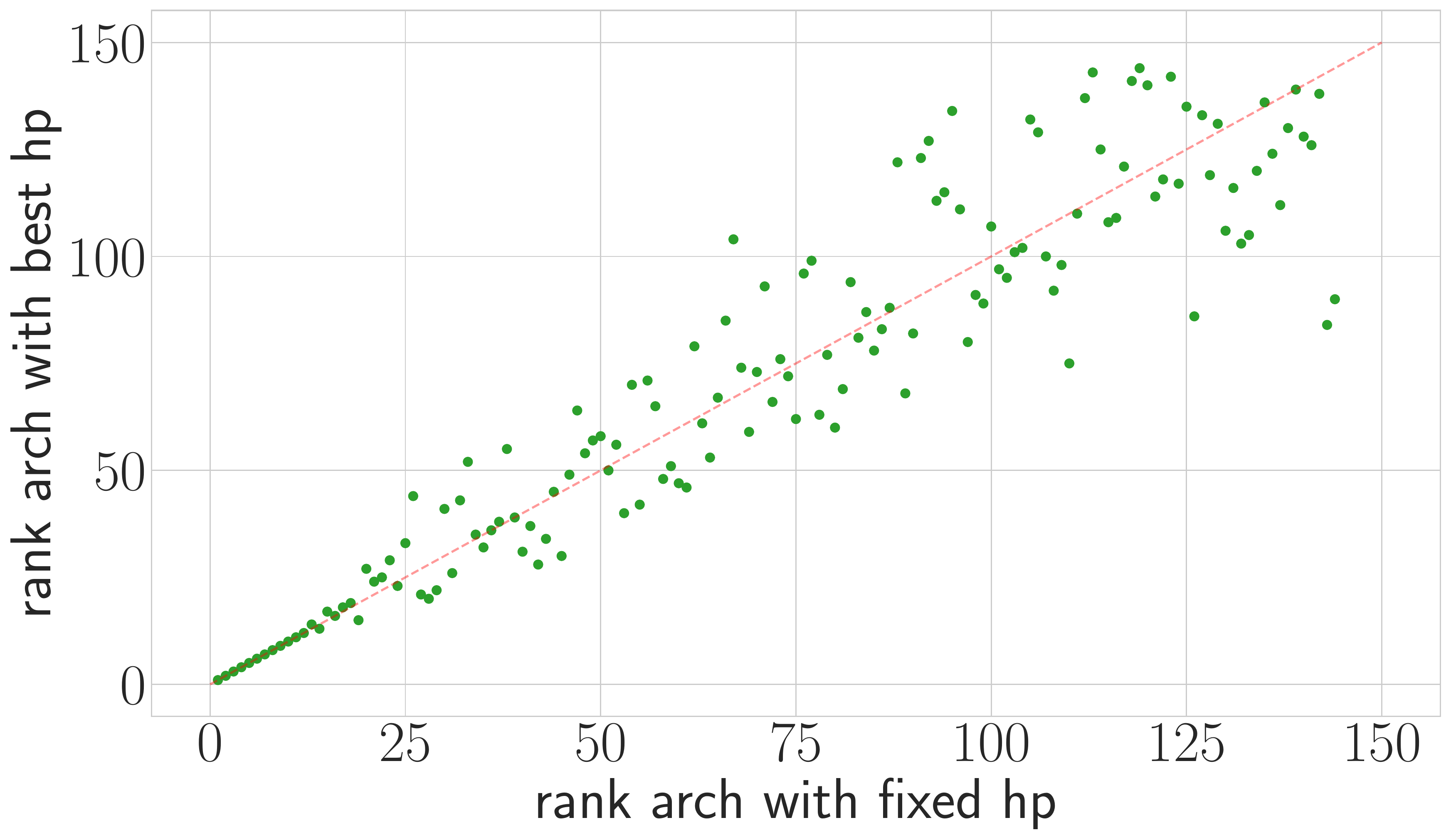}
\caption{Scatter plot between ranks of $f^*$ (vertical axis) and $f^\dagger$ (horizontal axis) on the \hpobench-Protein dataset. Ideally, the points should be close to the diagonal. The high correlation at low-rank means the best architectures are ranked identically when using $H^*$ and $H^\dagger$.}
 \label{fig:ranking_hpo_arch}
\end{center}
\end{figure}

All models in \nasbench were trained with a fixed set of hyperparameters. In this section, we justify that choice. The problem of hyperparameter optimization (HPO) is closely intertwined with NAS. NAS aims to discover good neural network architectures while HPO involves finding the best set of training hyperparameters for a given architecture. HPO operates by tuning various numerical neural network training parameters (\eg, learning rate) as well as categorical choices (\eg, optimizer type) to optimize the training process. Formally, given an architecture $A$, the task of HPO is to find its optimal hyperparameter configuration $H^*$:
$$H^*(A) = \argmax_H f(A, H),$$
where $f$ is a performance metric, such as validation accuracy and the $\argmax$ is over all possible hyperparameter configurations. The ``pure'' NAS problem can be formulated as finding an architecture $A^*$ when all architectures are evaluated under optimal hyperparameter choices:
$$A^* = \argmax_A f(A, H^*(A)),$$
In practice, this would involve running an inner HPO search for each architecture, which is computationally intractable. We therefore approximate $A^*$ with $A^\dagger$:
$$A^* \approx A^\dagger = \argmax_A f(A, H^\dagger),$$
where $H^\dagger$ is a set of hyperparameters that has been estimated by maximizing the average accuracy on a small subset $S$ of the architectures:
$$H^\dagger(S) = \argmax_H \overline{\{f(A, H) : A \in S\}}.$$
For example, in Section~\ref{sec:dataset_training}, $S$ was a random sample of 50 architectures.

To justify the approximation above, we performed a study
on a different set of \hpobench~\cite{hpobench} datasets (described in detail in  Supplement~\ref{sup:nas_hpo_bench})
These are smaller datasets of architecture--hyperparameter pairs $(A, H)$,
where we computed $f(A,H)$ for all settings of $A$ and $H$.
This let us compute the \textit{exact} hyperparameter-optimized accuracy,
$f^*(A) = \max_H f(A,H).$
We can also measure how well this correlates with the approximation we use in \nasbench.
To do this, we chose a set of hyperparameters $H^\dagger$ by optimizing the mean accuracy across all of the architectures for a given dataset. This allows us to map each architecture $A$ to its \textit{approximate} hyperparameter-optimized accuracy,
$f^\dagger(A) = f(A,H^\dagger).$
(This approximate accuracy is analogous to the one computed in the \nasbench metrics,
except there the average was over 50 random architectures, not all of them.)

We find that $f^\dagger$ and $f^*$ are quite strongly correlated across models, with a Spearman rank correlation of $0.9155$;
Figure~\ref{fig:ranking_hpo_arch} provides a scatter plot of $f^*$ against $f^\dagger$ for the architectures. The ranking is especially consistent for the best architectures (points near the origin).

\subsection{Absolute accuracy of models}

The choice of search space, hyperparameters, and training techniques were designed to ensure that \nasbench would be feasible to compute with our resources. Unfortunately, this means that the models we evaluate do not reach current state-of-the-art performance on CIFAR-10. This is primarily because: (1) the search space is constrained in both size and selection of operations and does not contain more complex architectures, such as those used by NASNet \cite{zoph2017learning}; (2) We do not apply the expensive ``augmentation trick'' \cite{zoph2017learning} by which models' depth and width are increased by a large amount and the training lengthened to hundreds of epochs; and (3) we do not utilize more advanced regularization like Cutout \cite{devries2017improved}, ScheduledDropPath \cite{zoph2017learning} and decoupled weight decay~\cite{loshchilov2019decoupled} in order to keep our training pipeline similar to previous standardized models like ResNet.

\section{Conclusion}

We introduced \nasbench, a new tabular benchmark for neural architecture search that is inexpensive to evaluate but still preserves the original NAS optimization problem, enabling us to rigorously compare various algorithms quickly and without the enormous computational budgets often used by projects in the field.
Based on the data we generated for this dataset, we were able to analyze the properties of an exhaustively evaluated set of convolutional neural architectures at unprecedented scale.
In open-sourcing the \nasbench data and generating code, we hope to make NAS research more accessible and reproducible.
We also hope that \nasbench will be the first of a continually improving sequence of rigorous benchmarks for the emerging NAS field.

\section*{Acknowledgements}
Aaron and Frank gratefully acknowledge support by the European Research Council (ERC) under the European Union's Horizon 2020 research and innovation programme under grant no.~716721, by BMBF grant DeToL, by the state of Baden-W\"{u}rttemberg through bwHPC and the German Research Foundation (DFG) through grant no INST 39/963-1 FUGG, and by a Google Faculty Research Award.
Chris, Esteban, Eric, and Kevin would like to thank Quoc Le, Samy Bengio, Alok Aggarwal, Barret Zoph, Jon Shlens, Christian Szegedy, Jascha Sohl-Dickstein; and the larger Google Brain team.

\bibliography{nasbench_paper}
\bibliographystyle{icml2019}

\FloatBarrier
\clearpage
\renewcommand{\figurename}{Supplementary Figure}
\renewcommand{\tablename}{Supplementary Table}
\renewcommand{\thesection}{S\arabic{section}}
\setcounter{section}{0}
\setcounter{figure}{0}
\setcounter{table}{0}
\twocolumn[
\icmltitle{\nasbenchnoformat: Towards Reproducible Neural Architecture Search}
\centerline{\textbf{\LARGE Supplementary Material}}
\vspace{30pt}
]

\section{Identifying Isomorphic Cells}
\label{sup:deduping_supplement}

\begin{figure}[ht]
    \includegraphics[width=\linewidth]{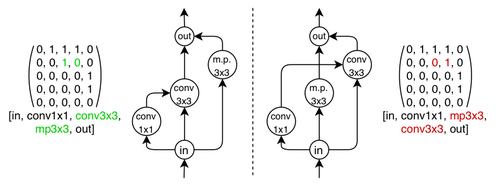}
    \caption{Two cells which are represented differently according to their adjacency matrix and labels but encode the same computation.}
    \label{isomorphic_cells}
\end{figure}

Within the \nasbench search space of models, there are models which have different adjacency matrices or have different labels but are computationally equivalent (e.g., Figure \ref{isomorphic_cells}).
We call such cells \textit{isomorphic}.
Furthermore, vertices not on a path from the input vertex to the output vertex do not contribute to the computation of the cell.
Cells with such vertices can be pruned to smaller cell without changing the effective behavior of the cell in the network.
Due to the size of the search space, it would be computationally intractable (and wasteful) to evaluate each possible graph representation without considering isomorphism.

Thus, we utilize an iterative graph hashing algorithm, described in \cite{ying2019enumerating}, which quickly determines whether two cells are isomorphic.
To summarize the algorithm, we iteratively perform isomorphism-invariant operations on the vertices of the graph which incorporates information from both the adjacent vertices as well as the vertex label.
The algorithm outputs a fixed-length hash which uniquely identifies isomorphic cells (i.e., computationally identical graphs cells to the same value and computationally different cells hash to different values).

Using such an algorithm allows us to enumerate all unique cells within the space and choose a single canonical cell to represent each equivalence class of cells and perform the expensive train and evaluation procedure on the canonical cell only.
When querying the dataset for a valid model, we first hash the proposed cell then use the hash to return the data associated with the evaluated canonical graph.

\section{Implementation Details}
\label{sup:implementation_details}

\subsection{Generating the dataset}
Table~\ref{tab:hyperparameters} shows the training hyperparameters used for all models in the space. These values were tuned to be optimal for the average of 50 randomly sampled cells in the search space. In practice, we find that these hyperparameters do not significantly affect the ranking of cells as long as they are set within reasonable ranges.

\begin{table}
\begin{tabular}{|l|l|}
    \hline
    batch size & 256 \\
    initial convolution filters & 128 \\
    learning rate schedule & cosine decay \\
    initial learning rate & 0.2 \\
    ending learning rate & 0.0 \\
    optimizer & RMSProp \\
    momentum & 0.9 \\
    L2 weight decay & 0.0001 \\
    batch normalization momentum & 0.997 \\
    batch normalization epsilon & 0.00001 \\
    accelerator & TPU v2 chip \\
    \hline
\end{tabular}
\caption{Important training hyperparamters.}
\label{tab:hyperparameters}
\end{table}

\subsection{Benchmarked algorithms}\label{sec:supp_hyperopt}

All methods employ the same encoding structure as defined in Section \ref{sec:encoding}.
For each method except random search, which is parameterfree, we identified the method's key hyperparameters and found a well-performing setting by a simple grid search which follows the same experimental protocol as described in the main text. Scripts to reproduce our experiments can be found at
\href{https://github.com/automl/nas_benchmarks}{https://github.com/automl/nas\_benchmarks}.

\paragraph{Random search (RS)}
We used our own implementation of random search which samples architectures simply from a uniform distribution over all possible configurations in the configuration space.

\begin{figure*}[h!]
\begin{center}
 \includegraphics[width=0.49\linewidth]{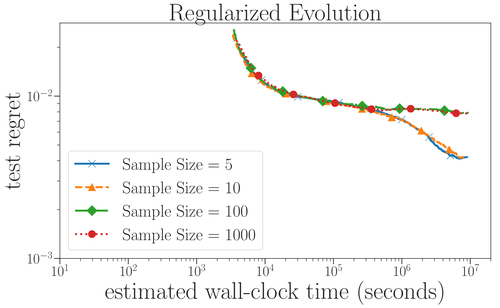}
 \includegraphics[width=0.49\linewidth]{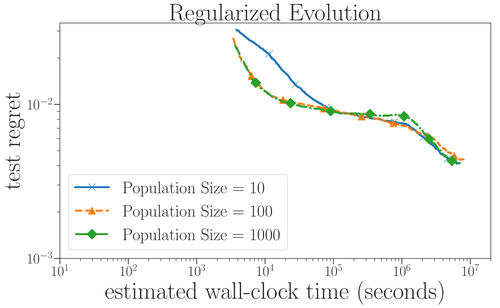}
  \caption{Meta parameters of RE. Left: Tournament Size; Right: Population Size.}
 \label{fig:hpo_re}
\end{center}
\end{figure*}

\paragraph{Regularized evolution (RE)}
We used a publicly available re-implementation for RE~\citep{real2018regularized}.
To mutate an architecture, we first sample uniformly at random an edge or an operator. If we sampled an edge we simply flip it and for operators, we sample a new operator for the set of all possible operations excluding the current one. RE kills the oldest member of the population at each iteration after reaching the population size.
We evaluated different values for the population size (PS) and the tournament size (TS) (see Figure~\ref{fig:hpo_re}) and set them to PS=100 and TS=10 for the final evaluation.

\paragraph{Tree-structured Parzen estimator (TPE)}
We used the Hyperopt implementation from \url{https://github.com/hyperopt/hyperopt} for TPE. All hyperparameters were left to their defaults, since the open-source implementation does not expose them and, hence, we could not adapt them for the comparison.

\paragraph{Hyperband}

For Hyperband we used the publicly available implementation from \url{https://github.com/automl/HpBandSter}. We set $\eta$ to 3 which is also used in \citet{li2017hyperband} and \citet{falkner2018bohb}. Note that, changing $\eta$ will lead to different budgets, which are not included in \nasbench.

\begin{figure*}[h!]
\begin{center}
 \includegraphics[width=0.24\linewidth]{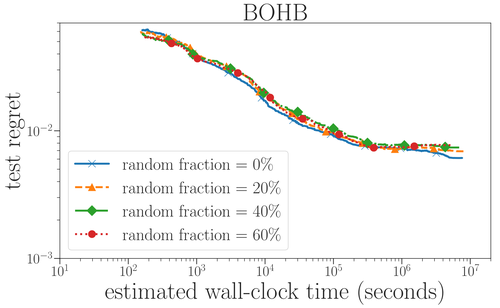}
 \includegraphics[width=0.24\linewidth]{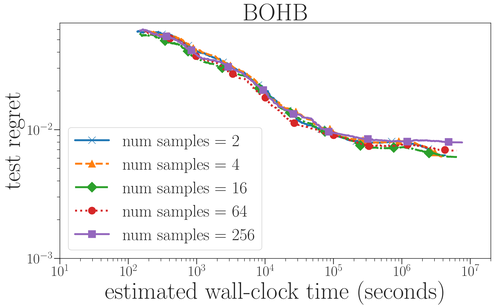}
 \includegraphics[width=0.24\linewidth]{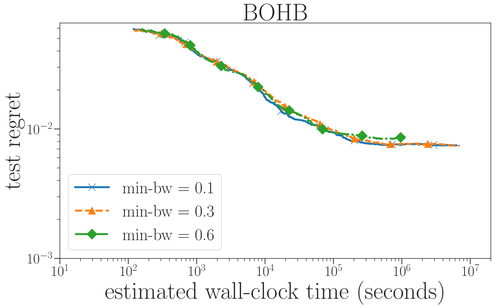}
 \includegraphics[width=0.24\linewidth]{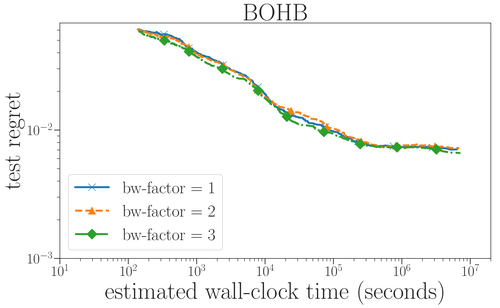}
  \caption{Performance of different meta parameters of BOHB. Left: fraction of random architectures; Middle Left: number of samples to optimize the acquisition function; Middle Right: minimum allowed bandwidth of the kernel density estimator; Right: Factor that is multiplied on the bandwidth for exploration.}
 \label{fig:hpo_bohb}
\end{center}
\end{figure*}

\paragraph{BOHB}
For BOHB we also used the implementation from \url{https://github.com/automl/HpBandSter}.
Figure~\ref{fig:hpo_bohb} shows the performance of different values for the fraction of random configurations, the number of samples to optimize the acquisition function, the minimum allowed bandwidth for the kernel density estimator and the factor which is multiplied to the bandwidth. 
Interestingly, while the minimum bandwidth and the bandwidth-factor do not seem to have an influence, the other parameters help to improve BOHB's performance, especially at the end of the optimization, if they are set to quite aggressive values.
For the final evaluation we set the random fraction to $0\%$, the number of samples to 4, the minimum-bandwidth to 0.3 (default) and the bandwidth factor to 3 (default).

\paragraph{Sequential model-based algorithm configuration (SMAC)}
We used the implementation from \url{https://github.com/automl/SMAC3} for SMAC.

\begin{figure*}[h!]
\begin{center}
 \includegraphics[width=0.32\linewidth]{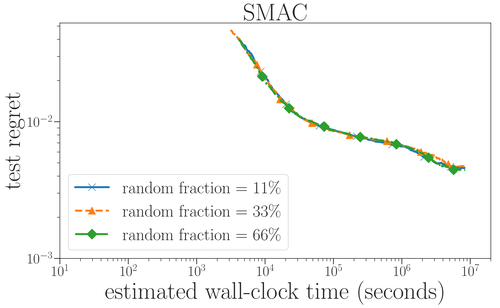}
 \includegraphics[width=0.32\linewidth]{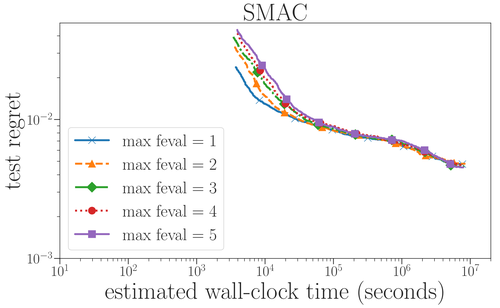}
 \includegraphics[width=0.32\linewidth]{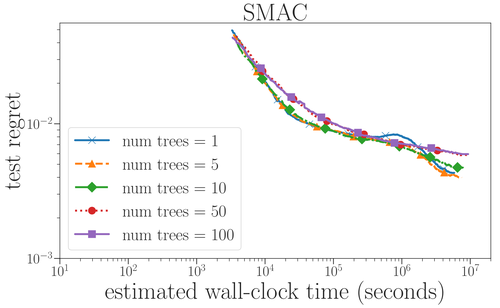}
  \caption{Performance of different meta parameters of SMAC. Left: fraction of random architectures; Middle: maximum number of function evaluations per architecture; Right: Number of trees in the random forest model.}
 \label{fig:hpo_smac}
\end{center}
\end{figure*}

As meta-parameters we exposed the fraction of random architecture that are evaluated, the maximum number of function evaluations per architecture and the number of trees of the random forest (see Figure~\ref{fig:hpo_smac}).
Since the fraction of random configurations does not seem to have an influence on the final performance of SMAC we kept it as its default (33$\%$). Interestingly, a smaller number of trees seems to help and we set it to 5 for the final evaluation. Allowing to evaluate the same configuration multiple times slows SMAC down in the beginning of the search, hence, we keep it at 1.

\paragraph{Reinforcement Learning}

\begin{figure}[h!]
\begin{center}
 \includegraphics[width=\linewidth]{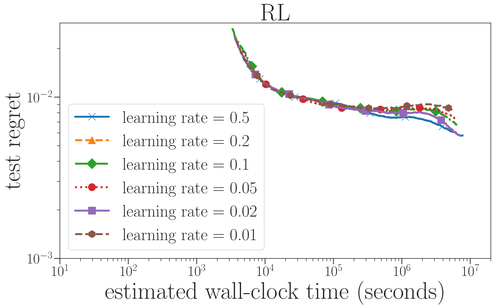}
  \caption{Right: Learning rate of our reinforcement learning agent.}
 \label{fig:hpo_rl}
\end{center}
\end{figure}

Figure~\ref{fig:hpo_rl} right shows the effect of the learning rate for our reinforcement learning agent described in Section~\ref{sec:rl}. For the final evaluation we used a learning rate of 0.5.

\section{Encoding}\label{sec:supp_encoding}

Besides the encoding described in Section~\ref{sec:nas-benchmark}, we also tried another encoding of the architecture space, which implicitly contains the constraint of a maximum of 9 edges.
Instead of having a binary vector for all the 21 possible edges in our graph, we defined for each edge $i$ a numerical parameter in $p_i \in [0, 1]$.
Additionally, we defined an integer parameter $N \in {0, ..., 9}$.
Now, in order to generate an architecture, we pick the $N$ edges with the highest values.
The encoding for the operators stays the same.

The advantage of this encoding is that by design no architecture violates the maximum number of edges constraint.
The major disadvantage is that some methods, such as regularized evolution or reinforcement learning, are not easily applicable without major changes due to the continuous nature of the search space.

\begin{figure}[h!]
\begin{center}
 \includegraphics[width=\linewidth]{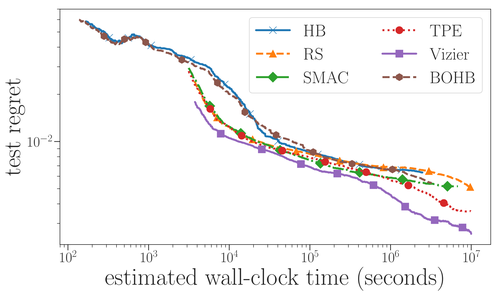}
  \caption{Comparison with a different encoding of architectures (see Section~\ref{sec:supp_encoding} for details). The experimental setup is the same as for Figure~\ref{fig:comparison} in the main text, but note that the hyperparameters of BOHB and SMAC were determined based on the main encoding and are not optimal for this encoding.}
 \label{fig:encoding}
\end{center}
\end{figure}

Figure~\ref{fig:encoding} shows the comparison of all the methods that can be trivially applied to this encoding.
We used the same setup as described in Section~\ref{sec:nas-benchmark}.
Additionally, we also include Vizier, which is not applicable to the default encoding.
All hyperparameters are the same as described in Section~\ref{sec:supp_hyperopt}.
Interestingly the ranking of algorithms changed compared to the results in Figure~\ref{fig:comparison}.
TPE achieves a much better performance now than on the default encoding and outperforms SMAC and BOHB.
We assume that, since we used the hyperparameters of SMAC and BOHB that were optimized for the default encoding in Section~\ref{sec:supp_hyperopt}, they do not translate to this new encoding.

\begin{figure}[t]
\begin{center}
  \includegraphics[width=0.98\linewidth]{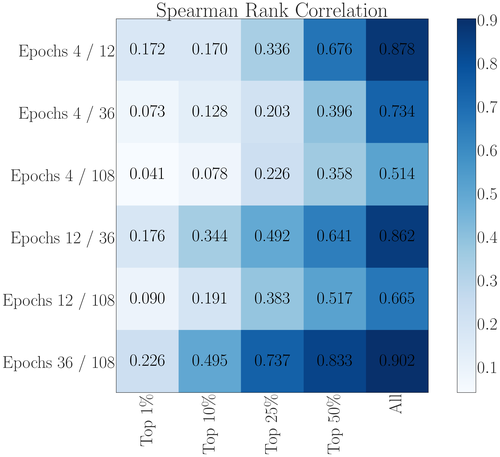}
  \caption{The Spearman rank correlation between accuracy at different number of epoch pairs (rows) for different percentiles of the top architectures (columns) in \nasbench. \textit{E.g.}, the accuracies between 36 and 108 epochs across the top-10\% of architectures have a 0.365 correlation.}
 \label{fig:rank_correlation}
\end{center}
\end{figure}

\section{REINFORCE Baseline Approach}\label{sec:rl}

We attempted to benchmark a reinforcement learning (RL) approach using a 1-layer LSTM controller trained with PPO, as proposed by \citet{zoph2017learning}. With no additional hyperparameter tuning, the controller seems to fail to learn to traverse the space and tends to converge quickly to a far-from-optimal configuration. We suspect that one reason for this is the highly conditional nature of the space (i.e., cells with more than 9 edges are "invalid"). Further tuning may be required to get RL techniques to work on \nasbench, and this constitutes an interesting direction for future work.

\begin{figure}[t]
\begin{center}
  \includegraphics[width=0.98\linewidth]{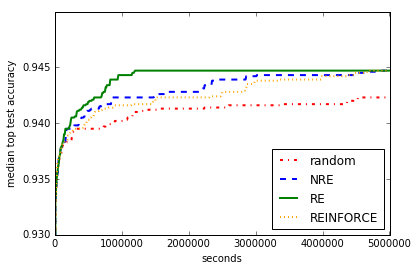}
  \caption{Comparing REINFORCE against regularized evolution (RE), non-regularized evolution (NRE), and a random search baseline (RS).}
 \label{fig:reinforce compare}
\end{center}
\end{figure}

We did, however, successfully train a naive REINFORCE-based \cite{DBLP:journals/ml/Williams92} controller which simply outputs a multinomial probability distribution at each of the 21 possible edges and 5 operations and samples the distribution to get a new model. We believe that this sampling behavior allows it to find more diverse models than the LSTM-PPO method. The results, when run in the same context as Section \ref{sec:analysis_generalization}, are shown in Figure \ref{fig:reinforce compare}. REINFORCE appears to perform around as strongly as non-regularized evolution (NRE) but both NRE and REINFORCE tends to be weaker than regularized evolution (RE). All methods beat the baseline random search.

 \section{The \hpobench Datasets}
\label{sup:nas_hpo_bench}

The \hpobench datasets consists of 62208 hyperparameter configurations of a 2-layer feedforward networks on four different non-image regression domains, making them complementary to \nasbench. 
We varied the number of hidden units, activation types and dropout in each layer as well as the learning rate, batch size and learning rate schedule.
While the graph space is much smaller than \nasbench, it has the important advantage of including hyperparameter choices in the search space, allowing us to measure their interaction and relative importance.
For a full description of these datasets, we refer to \citet{hpobench}.

\section{Guidelines for Future Benchmarking of Experiments on NAS-Bench-101}\label{sec:guidelines}
To facilitate a standardized use of NAS-Bench-101 in the future benchmarking of algorithms, we recommend the following practices:
\begin{enumerate}
\item Perform many runs of the various NAS algorithms (in our experiments, we ran 500).
\item  Plot performance as a function of estimated wall clock time and/or number of function evaluations (as in our Figure \ref{fig:comparison}, left). This allows judging the performance of algorithms under different resource constraints. To allow this, every benchmarked algorithm needs to keep track of the best architecture found up to each time step.
\item  Do not use test set error during the architecture search process. In particular, the choice of the best architecture found
up to each time step can only be based on the training and validation sets. The test error can only be used for offline evaluation once the search runs are complete.
\item  To assess robustness of the algorithms with respect to the seed of the random number generator, plot the empirical cumulative distribution of the many runs performed; see our Figure \ref{fig:comparison} (right) for an example.
\item  Compare algorithms using the same hyperparameter settings for NAS-Bench-101 as for other benchmarks. Even though tabular benchmarks like NAS-Bench-101 allow for cheap comprehensive evaluations of different hyperparameter settings (see the next point), in practice NAS algorithms need to come with a set of defaults that the authors propose to use for new NAS benchmarks (or an automated/adaptive method for setting the hyperparameters online); the performance of these defaults should be evaluated. 
\item  Report performance with different hyperparameter settings to produce a quantitative sensitivity analysis (as in Figures \ref{fig:hpo_smac}-\ref{fig:hpo_rl} of this appendix).
\item  If applicable, also study performance for alternative encodings, such as the continuous encoding discussed in Appendix \ref{sec:supp_encoding}.
\end{enumerate}

\end{document}